\documentclass[letterpaper, 10 pt, conference]{ieeeconf}  

\IEEEoverridecommandlockouts                              

\usepackage{graphicx}
\usepackage{color}
\usepackage{cite}
\usepackage[hidelinks,bookmarks=false]{hyperref}
\usepackage{multirow}
\usepackage{enumerate}
\usepackage{bm}
\usepackage{booktabs}
\usepackage{blindtext}
\usepackage[T1]{fontenc}
\usepackage{mathptmx} 
\DeclareMathAlphabet{\mathcal}{OMS}{cmsy}{m}{n}
\usepackage{times} 
\usepackage{amsmath} 
\usepackage{amssymb}  
\usepackage[ruled,linesnumbered]{algorithm2e}

\usepackage{hyperref}
\hypersetup{
hidelinks,
colorlinks=true,
linkcolor=red,
citecolor=green,
filecolor=magenta,
urlcolor=blue
}
\usepackage{xcolor}

\usepackage{bbm}
\usepackage{booktabs}
\usepackage{upgreek}
\usepackage{mathtools,xparse}
\usepackage{colortbl}

\newcommand{\PreserveBackslash}[1]{\let\temp=\\#1\let\\=\temp}
\newcolumntype{C}[1]{>{\PreserveBackslash\centering}p{#1}}
\newcolumntype{R}[1]{>{\PreserveBackslash\raggedleft}p{#1}}
\newcolumntype{L}[1]{>{\PreserveBackslash\raggedright}p{#1}}

\DeclarePairedDelimiter{\norm}{\lVert}{\rVert}
\NewDocumentCommand{\normL}{ s O{} m }{%
  \IfBooleanTF{#1}{\norm*{#3}}{\norm[#2]{#3}}_{L_2($\Omega$)}%
}

\DeclareFontFamily{U} {cmmi}{}

\DeclareFontShape{U}{cmmi}{m}{n}{
  <-6> cmmi5
  <6-7> cmmi6
  <7-8> cmmi7
  <8-9> cmmi8
  <9-10> cmmi9
  <10-12> cmmi10
  <12-> cmmi12}{}

\DeclareSymbolFont{Xcmmi} {U} {cmmi}{m}{n}

\DeclareMathSymbol{\psi}{\mathord}{Xcmmi}{32}

\makeatletter

\newcommand{\Rmnum}[1]{\expandafter\@slowromancap\romannumeral #1@}
\makeatother

\begin{document}
\title{Carl-Lead: Lidar-based End-to-End Autonomous Driving with\\Contrastive Deep Reinforcement Learning}

\author{Peide~Cai,
        Sukai~Wang,
        Hengli~Wang,
        and~Ming~Liu,~\IEEEmembership{Senior Member,~IEEE}
\thanks{This work was supported by Zhongshan Municipal Science and Technology Bureau Fund, under project ZSST21EG06, Collaborative Research Fund by Research Grants Council Hong Kong, under Project No. C4063-18G, and Department of Science and Technology of Guangdong Province Fund, under Project No. GDST20EG54, awarded to Prof. Ming Liu. \textit{(Corresponding author: Ming Liu.)}}
\thanks{All authors are with the Hong Kong University of Science and Technology, Clear Water Bay, Kowloon, Hong Kong SAR, China (e-mail: \{pcaiaa, hwangdf, swangcy\}@connect.ust.hk, eelium@ust.hk) }
}

\maketitle

\begin{abstract}

Autonomous driving in urban crowds at unregulated intersections is challenging, where dynamic occlusions and uncertain behaviors of other vehicles should be carefully considered. Traditional methods are heuristic and based on hand-engineered rules and parameters, but scale poorly in new situations. Therefore, they require high labor cost to design and maintain rules in all foreseeable scenarios. Recently, deep reinforcement learning (DRL) has shown promising results in urban driving scenarios. However, DRL is known to be sample inefficient, and most previous works assume perfect observations such as ground-truth locations and motions of vehicles without considering noises and occlusions, which might be a too strong assumption for policy deployment. In this work, we use DRL to train lidar-based end-to-end driving policies that naturally consider imperfect partial observations. We further use unsupervised contrastive representation learning as an auxiliary task to improve the sample efficiency. The comparative evaluation results reveal that our method achieves higher success rates than the state-of-the-art (SOTA) lidar-based end-to-end driving network, better trades off safety and efficiency than the carefully-tuned rule-based method, and generalizes better to new scenarios than the baselines. Demo videos are available at \url{https://caipeide.github.io/carl-lead/}.

\end{abstract}

\section{Introduction}

\begin{figure}[t]
        \centering
        \includegraphics[width = 0.9\columnwidth]{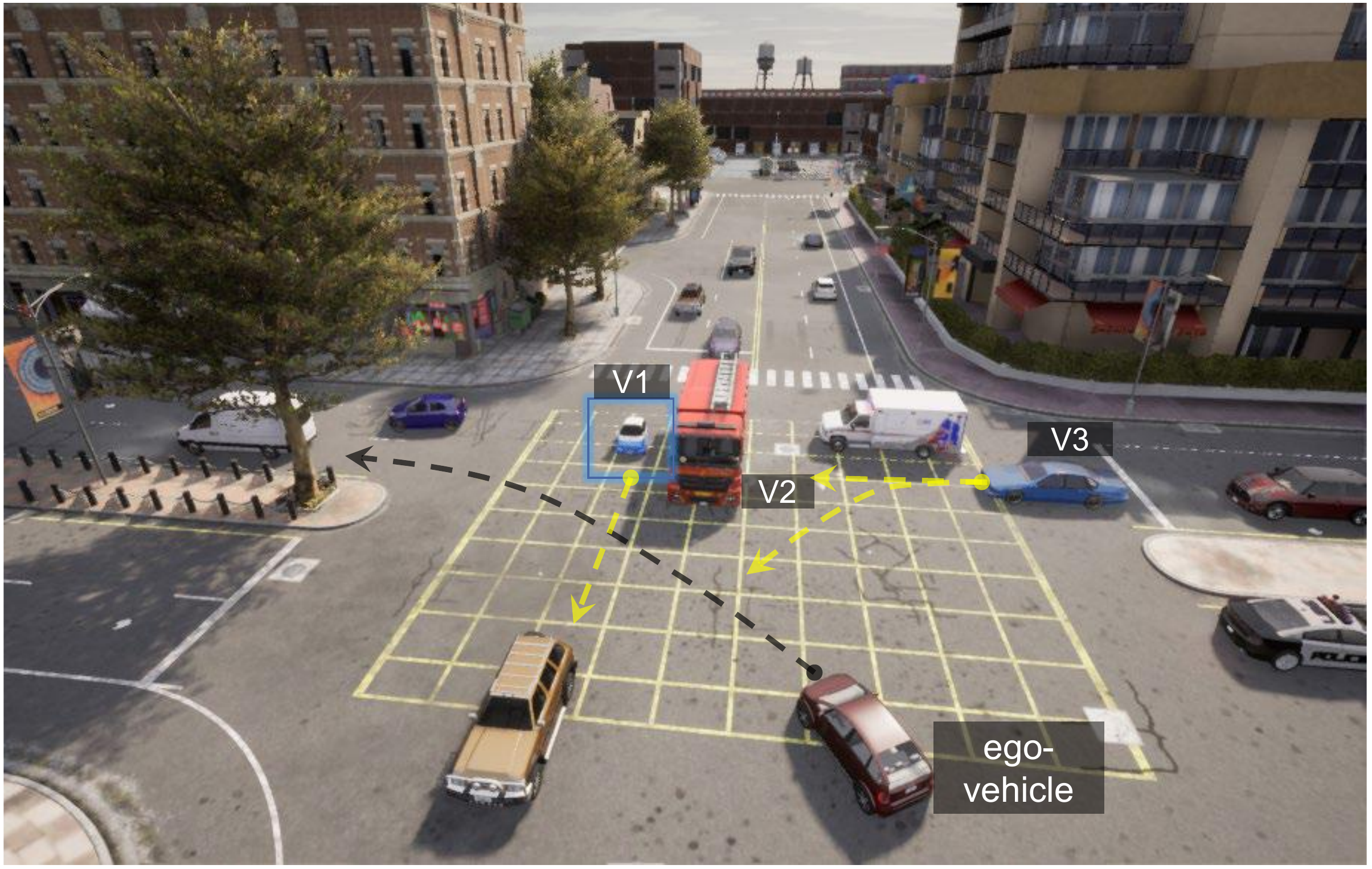}
        \caption{A driving scenario considered in this work, where the ego-vehicle should take an unprotected left turn at the unsignalized multi-lane intersection. However, the opposite red truck V2 occludes the lateral small mini-car V1. Therefore, the ego-vehicle may neglect V1 and collide with it. In addition, the behaviors of different vehicles are hard to model manually. For example, vehicle V3 may plan to go straight or turn left, in an aggressive or a mild driving pattern. In this work, we use DRL to learn end-to-end driving policies for these uncertain and interactive environments.}
        \label{fig:motivation}
\end{figure}

Autonomous driving aims to improve the safety and efficiency of our daily life, and research on this topic has significantly developed in the last decade\cite{cosgun2017towards, kuutti2020survey, tampuu2020survey}. However, driving in  unregulated urban environments with other road vehicles, like unsignalized intersections (Fig. \ref{fig:motivation}), remains an open problem. The main difficulties are twofold: 1) the behavioral patterns (e.g., timid, moderate and aggressive) and intents of other vehicles are complex and cannot be directly observed, and 2) the perception of the autonomous vehicle (AV) is uncertain due to noises and occlusions. 

Traditional planning methods for AVs are based on manually-designed rules, such as the finite state machines (FSM) adopted in the DARPA Urban Challenge\cite{montemerlo2008junior} and recent applications\cite{cosgun2017towards}. However, these methods are hard to tune for highly complex environments considering uncertainties and interactions\cite{lin2019decision}. In addition, they have difficulties to generalize to new scenarios\cite{kuutti2020survey}, and may suffer from the detection errors from the upper-stream perception module, such as the unfortunate Tesla accident in 2016\cite{tesla}.

The above challenges have been motivating recent works to use deep learning for decision making of AVs, where the driving policies can be learned and optimized from large amount of data without hand-crafted rules. A growing research trend is called end-to-end driving\cite{cai2020probabilistic, Cai2020VTGNetAV, pan2020imitation, liu2021efficient, codevilla2018end, gao2017intention, Pfeiffer2017FromPT, cai2021vision, kendall2019learning}, which aims to directly map the raw sensory data (e.g., RGB images) to driving commands with deep neural networks. These networks can be either trained with imitation learning (IL) using supervised data\cite{cai2020probabilistic, Cai2020VTGNetAV, pan2020imitation, liu2021efficient, codevilla2018end, gao2017intention, Pfeiffer2017FromPT}, or with deep reinforcement learning (DRL) using trial-and-error explorations\cite{cai2021vision, kendall2019learning}. However, IL suffers from the distribution shift problem\cite{tampuu2020survey}, and DRL is known to be sample inefficient dealing with high-dimensional data. For more efficient training and better results, some works directly use the low-dimensional but fully observed (ground-truth) state information of traffic environments, such as locations and motions of other vehicles, as input to the deep networks\cite{chen2019model, cai2021dignet, saxena2020driving}. However, this further brings a problem of neglecting the perception noises and occlusions, leaving it unclear if these methods are suitable for deployment. In this work, we present Carl-Lead, a \underline{L}idar-based \underline{e}nd-to-end \underline{a}utonomous \underline{d}riving method trained with \underline{C}ontr\underline{a}stive deep \underline{r}einforcement \underline{l}earning. The key contributions are summarized as follows:

\begin{figure*}[t]
        \centering
        \includegraphics[width = 2\columnwidth]{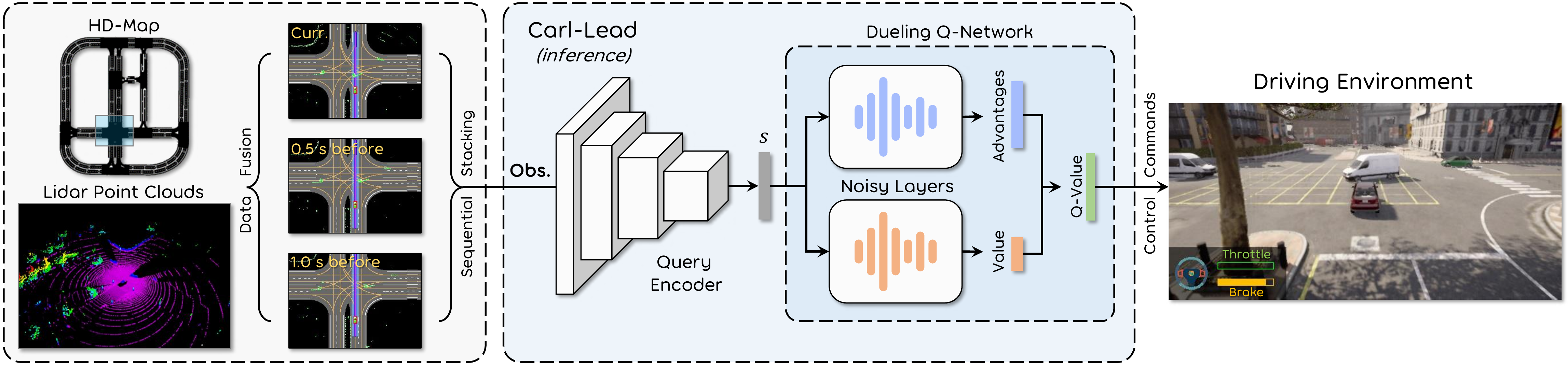}
        \caption{Overview of the \textit{inference pipeline} for our proposed lidar-based end-to-end autonomous driving network, Carl-Lead.}
        \label{fig:network}
\end{figure*}

\begin{enumerate}
    
    \item We propose an end-to-end autonomous driving method for navigating unregulated urban intersections using raw lidar point clouds (as shown in Fig. \ref{fig:network}), capable of handling imperfect partial observations such as occlusions.
    
    \item We use the model-free, off-policy DRL to train the agent in different traffic scenarios, so as to understand and adapt to the uncertain behaviors of other vehicles in an interactive manner for safe and efficient driving.
    
    \item We experimentally demonstrate that unsupervised contrastive learning can improve both training efficiency and driving performance in our task.
    
    \item We systematically evaluate different methods, showing that our approach achieves higher success rates than the learning-based baselines, better trades off safety and efficiency than the rule-based method, and generalizes better in new environments (e.g., roundabouts).
    
\end{enumerate}

\section{Related Work}
\textbf{End-to-end autonomous driving.} In recent years, a new deep learning-based pipeline for AVs integrating perception, planning and control, named end-to-end driving, has achieved impressive results\cite{cai2020probabilistic, Cai2020VTGNetAV, cai2021vision, pan2020imitation, liu2021efficient, codevilla2018end, kendall2019learning, gao2017intention, Pfeiffer2017FromPT}. In this area, IL has been the most popular method for network training due to its high sample efficiency, where the model tries to mimic the observation-action data pairs collected by the expert. For example, Codevilla \textit{et al.}\cite{codevilla2018end} designed a conditional end-to-end network to take as input both camera images and high-level commands (e.g., \textit{keep straight}, \textit{turn left}) for autonomous driving. More recently, Liu \textit{et al.}\cite{liu2021efficient} developed an uncertainty-aware lidar-based driving network, since lidar provides more accurate 3D information and greater robustness to illumination changes than the camera. However, IL suffers from the distribution shift problem\cite{tampuu2020survey}, which can lead to the model making mistakes in new situations different from the distribution of the training set. 

Another training paradigm is DRL, with which the agent can automatically learn knowledge by interacting online with the environment in a trial-and-error manner, without relying on supervisions. However, DRL is much less data efficient than IL. For example, Kalashnikov \textit{et al.}\cite{kalashnikov2018qt} used multiple robotic arms for collecting interaction data over \textit{four months} to train their vision-based robot grasping policies. Therefore, DRL has only reached limited results in the form of end-to-end driving, such as lane following\cite{kendall2019learning, cai2021vision} without considering dynamic urban scenarios. In this work, we use DRL to train lidar-based end-to-end networks, but differently, we consider more complex and interactive urban driving scenarios, and use the recent unsupervised contrastive representation learning\cite{laskin2020curl, oord2018representation} to improve the sample efficiency of DRL in our task (Sec. \ref{subsec: drl}).

\textbf{Autonomous driving at intersections.} Traditional methods in the automotive industry are based on hand-crafted rules, and the time-to-collision (TTC)\cite{minderhoud2001extended} is a widely used heuristic safety indicator in FSMs for intersection handling of AVs\cite{isele2018navigating, montemerlo2008junior, cosgun2017towards}. While TTC is relatively reliable and easy to interpret, it has many limitations: 1) It assumes constant velocity, but the states of other vehicles are dynamically changing in traffic. This oversimplified assumption may lead to errors for safety measurement and further cause accidents. 2) TTC-based methods are reactive, meaning they tend to slow down and wait other vehicles in case of collision, often generating overly-conservative bahaviors.

Due to the above limitations, recent works have been trying to use DRL techniques for intersection handling problems\cite{isele2018navigating, chen2019model}, where the agent can proactively interact with other traffic participants rather than behave reactively. However, these works either assume fully observed traffic information\cite{chen2019model}, or perfect perception and state estimation\cite{isele2018navigating} of surroundings without noises. These strong assumptions may not be applicable for applications. To naturally consider imperfect perceptions for autonomous driving, in this work we use the raw lidar data as partial observations of the environments, and use DRL for training end-to-end driving policies to adapt to the occlusions and noises in perception.

\section{Methodology}
\label{sec: method}

\subsection{Deep Reinforcement Learning}
\label{subsec: drl}
\subsubsection{Problem Formulation}

We first define the target autonomous driving problem as a Markov decision process (MDP), which can be formulated as a tuple containing five elements: $<\mathcal{S}, \mathcal{A}, R, f, \gamma>$, which denote the state space, action space, immediate reward, state transition model and the discount factor, respectively. Within this formulation, the agent interacts with environment and learns the policy $\pi$ by maximizing the expected discounted return $R_t = \sum_{\tau=t}^{\infty}\gamma^{\tau-t}r_{\tau}$.

\subsubsection{Deep Q-Learning}
Given a policy $\pi$, the action-value (Q) is defined as the expected return for a state-action pair:
\begin{equation}
Q_{\pi}(s_t, a_t)=\mathbb{E}_{\pi}\left[R_{t} \mid s_t, a_t\right].
\end{equation}
In deep Q-learning, the optimal value function is approximated by deep neural networks with parameters $\theta$: $Q^{*}(s, a) \approx Q(s, a ; \theta)$. Then, we can choose the optimal action $a_t^{*}$ with the largest Q-value to execute at state $s_t$. For model training, we adopt the dueling double deep Q-learning (D3QN) algorithm\cite{wang2016dueling}, and optimize the following loss function at iteration $i$ based on the temporal-difference (TD) error:
\begin{equation}
\mathcal{L}_{rl}\left(\theta_{i}\right) =\mathbb{E}_{s, a, r, s^{\prime}}[\left(y_{i}-Q\left(s, a ; \theta_{i}\right)\right)^{2}],
\label{eq:rl-loss}
\end{equation}
\begin{equation}
y_{i}=r+\gamma \hat{Q}(s^{\prime}, \underset{a^{\prime}}{\arg \max } Q\left(s^{\prime}, a^{\prime} ; \theta_{i}\right) ; \theta^{-}),
\end{equation}
where $\hat{Q}$ denotes the \textit{target network} with paramters $\theta^-$, which are updated by copying the weights of $Q(s,a;\theta)$ every $T_N$ steps and freezed in other intervals. Within D3QN, two streams of sub-networks are built to compute the state value $V_{\pi}(s)$ and advantage functions $A_{\pi}(s, a)$ (vector of |$\mathcal{A}$|-dimensional) separately, as shown in Fig. \ref{fig:network}. These two branches are finally combined to compute the action values according to Eq. (\ref{eq:dueling}):
\begin{equation}
\begin{aligned}
Q(s, a ; &\theta)=V(s ; \theta)+(A(s, a ; \theta)-\frac{1}{|\mathcal{A}|} \sum_{a^{\prime}} A\left(s, a^{\prime} ; \theta\right)).
\label{eq:dueling}
\end{aligned}
\end{equation}

\subsubsection{Noisy Networks for Exploration}
\label{subsec: noisy-nets}

Classical DRL methods use $\epsilon$-greedy strategies to randomly perturb agent's policy and induce novel behaviors for exploration. However, it requires hyper parameter tuning, and the induced \textit{local} dithering perturbations are hard to generate diverse behaviors for efficient exploration. Therefore, we adopt the idea of \textit{Noisy Nets}\cite{fortunato2018noisy} in this work for exploration. It uses a noisy linear layer that combines a deterministic and noisy stream:
\begin{equation}
\bm{y}=(\bm{b}+\mathbf{W} \bm{x})+\left(\bm{b}_{\text {noisy }} \odot \epsilon^{b}+\left(\mathbf{W}_{\text {noisy }} \odot \epsilon^{w}\right) \bm{x}\right),
\end{equation}
where the parameters $\bm{b},\mathbf{W}, \bm{b}_{\text{noisy}}$ and $\mathbf{W}_{\text{noisy}}$ are learnable whereas $\epsilon^{b}$ and $\epsilon^{w}$ are random variables. In this way, the amount of noise injected in the network is tuned automatically by the RL algorithm, allowing state-conditioned and self-annealing exploration. During testing, $\bm{b}_{\text{noisy}}$ and $\mathbf{W}_{\text{noisy}}$ are set to zero for a stable policy deployment.

\begin{figure*}[t]
        \centering
        \includegraphics[width = 1.8\columnwidth]{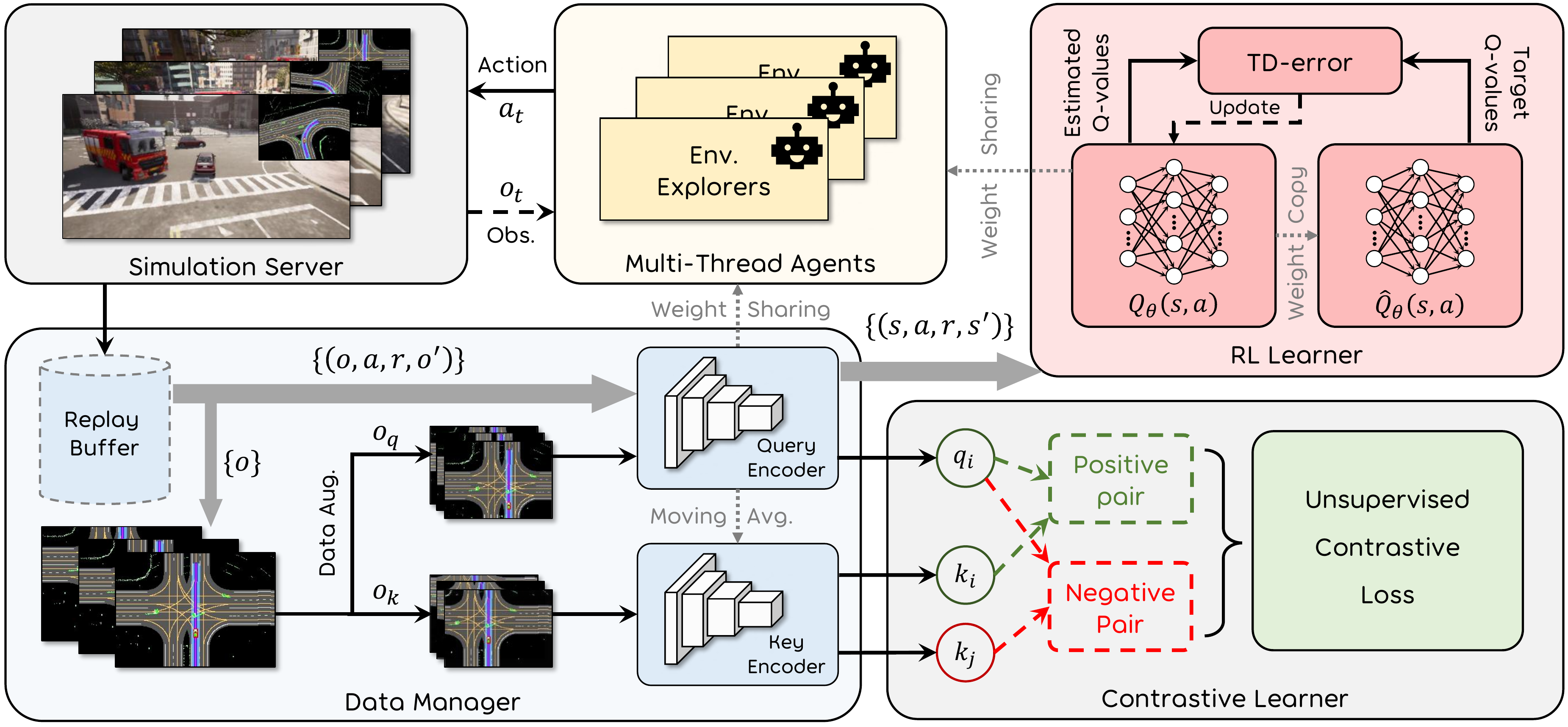}
        \caption{Overview of the \textit{training pipeline} for our method. In general, two kinds of loss functions are calculated: 1) the temporal difference (TD) loss of RL, and 2) the contrastive loss of the auxiliary representation learning process. The query encoder projects the original high-dimensional observations $o\in \mathbb{R}^{200\times280\times9}$ from the sampled transitions $\{(o, a, r, o^{\prime})\}$ to lower-dimensional latent state vectors $s\in\mathbb{R}^{512}$ for efficient RL training (the red block).}
        \label{fig:training}
\end{figure*}

\subsubsection{Contrastive Learning}
Contrastive learning \cite{oord2018representation} is an unsupervised framework to learn representations of high-dimensional data, and is shown to improve the sample efficiency as an concurrent auxiliary task of DRL in a series of Atari games\cite{laskin2020curl}. In this work, we extend this idea to the end-to-end autonomous driving domain, showing it is also beneficial to improve the final task performance (Sec. \ref{sec: experiments}). 

As shown in Fig. \ref{fig:training}, a batch of $N$ observations are first data-augmented (random cropping is used in this work) \textit{twice} to form \textit{query} and \textit{key} observations $o_q, o_k$, which are then projected into lower-dimensional representations $q_i, k_j \in \mathbb{R}^{512}$ ($1\leq i,j \leq N$) by the query and key encoders (ResNet18), respectively. A query vector $q_i$ and key $k_j$ is said to be positive pairs if they originate from the same image (i.e., i=j) and negative otherwise. Then, the contrastive representations are learned by training positive pairs to be more similar to each other, and the negative pairs to be dissimilar to each other. More specifically, the similarity of different query-key pairs are calculated with bi-linear inner-product: $\text{sim}(q,k) = q^TWk$, where $W$ is a learnable matrix $\in \mathbb{R}^{512\times512}$. Then we use the InfoNCE score function\cite{oord2018representation} coupled with this similarity measure to compute the contrastive loss to update the query encoder with parameters $\theta_q$:
\begin{equation}
\mathcal{L}_{c}=-\frac{1}{N} \sum_{i=0}^{N-1} \log \frac{\exp \left(\operatorname{sim}\left(q_{i}, k_{i}\right)\right)}{\sum_{j=0}^{N-1} \exp \left(\operatorname{sim}\left(q_{i}, k_{j}\right)\right)}.
\label{eq:contrastive-loss}
\end{equation}
The weights of the key encoder $\theta_k$ are the moving average of $\theta_q$ and updated at each training step as follows:
\begin{equation}
\theta_{k}=m \theta_{k}+(1-m) \theta_{q}.
\end{equation}

\subsection{Training Scenarios}
\label{subsec: scenarios}
We use the open-source CARLA simulator \cite{dosovitskiy2017carla} for training and evaluation of our method at different unregulated intersections, since it provides abundant vehicle models and traffic scenarios close to the real world. The training scenarios are: \texttt{T-Merge}, \texttt{Int-Cross}, \texttt{Int-Left}, and \texttt{T-Left}, shown in Fig. \ref{fig:qualitative}. For the scenarios, \texttt{T} means a T-shaped junction, and \texttt{Int} means a four-lane intersection. For driving tasks, \texttt{Merge} involves making a right turn for lane merging, \texttt{Left} indicates an unprotected left turn, and \texttt{Cross} means driving straight to cross the intersection. In order to cover as many driving occasions as possible, in each episode the properties of other vehicles are randomly initialized, including locations, types (cars, big trucks, ambulance, mini-cars, etc.), speeds, goal locations and so on. These vehicles are controlled by the AI engine of CARLA\footnote{{https://carla.readthedocs.io/en/0.9.12/adv\_traffic\_manager/}} for realistic traffic scenarios. The simulation step is set to 0.1 seconds, meaning the control frequency is 10 Hz for all our experiments.

\subsection{Implementation Details}
\subsubsection{State Space}
\label{subsec: state-space}
In each frame, we project the lidar point clouds into an occupancy grid map (OGM), which is then rendered on a HD-map image as information fusion. Using map information to learn driving policies is beneficial since it provides structural priors on the motion of other vehicles\cite{cai2021dignet}. In this work, our region of interest is 70 m wide (35 m to each side of the ego-vehicle) and 50 m long (35 m front and 15 m behind). The resolution is 0.25 m/pixel, resulting in an image of size ($200\times280\times3$) per frame. For the state space of the DRL algorithm, we stack a sequence of frame data as input observations $o_t$ to the network, aiming to maintain a temporal dependency and enable the network to implicitly model prediction and reasoning of the environment (occlusions, intents of other vehicles, and noises). The horizon is empirically set to 1.0 s with 3 frames, as shown in Fig. \ref{fig:network}.

\subsubsection{Action Space}
The action space of the ego-vehicle consists of its steering and speed controls. Specifically, we decouple this action space and make the agent compute the target speeds among $\mathcal{A} = \{0,10,20,30\}$ (km/h), which are translated to throttle/brake values by a PID controller to navigate the vehicle longitudinally. The lateral steering control is implemented with the pure-pursuit algorithm\cite{coulter1992implementation}.

\subsubsection{Reward Design}
To stimulate \textit{efficient} driving, the reward is set to ${v}/{30}$, which is proportional to the speed of the ego-vehicle and normalized to $[0, 1]$. If collision happens, the reward is set to -50 as punishment for \textit{safe} driving.

\subsubsection{Overall Pipeline}
Based on the components introduced in this section, the overview training pipeline of our Carl-Lead for autonomous driving is shown in Fig. \ref{fig:training}, which consists of five modules, i.e., \textbf{RL Learner}, \textbf{Agents} for exploration, \textbf{Simulation Server}, \textbf{Data Manager}, and \textbf{Contrastive Learner}. The training process is divided into two iterative parts until convergence: \textit{exploration} and \textit{model update}. At the exploration stage, multiple agents sharing the same network collect experiences in their separate CARLA threads, which are hosted in the simulation server. Transition tuples $e_{t}=\left(o_{t}, a_{t}, r_{t}, o_{t+1}\right)$ from each working thread are synchronized by the message passing interface (MPI) and stored in the replay buffer $\mathcal{D}$, enabling high throughput. At the training stage, mini-batches of data $\left\{e_1, e_2, ..., e_N\right\}$ are first sampled from $\mathcal{D}$, and are then processed by two branches: 1) the {RL branch} encodes the original observations $\{o\}$ from the batch data into latent states $\{s\}$ with the query encoder, to compute the RL loss $\mathcal{L}_{rl}$ based on Eq. (\ref{eq:rl-loss}); 2) the {contrastive learning branch} extracts the original observations $\{o\}$ from the batch data, and then compute the unsupervised contrastive loss $\mathcal{L}_{c}$ based on Eq. (\ref{eq:contrastive-loss}). Finally, both $\mathcal{L}_{rl}$ and $\mathcal{L}_{c}$ are jointly optimized through stochastic gradient decent.

\section{Experimental Setup}

\subsection{Training Setup}
To train our Carl-Lead, we run six CARLA simulators running in parallel for collecting experiences at scale in different scenarios. The size of the replay buffer is set to 400K and the batch size is 128. We use an NVIDIA RTX 3090 GPU to train the network end-to-end, and use the Adam optimizer with a learning rate of 0.0001. The discount factor $\gamma$ is 0.99, and the momentum weight $m$ to update the key encoder is 0.999. Finally, the training process costs 12 hours to converge with 488.2K interaction steps.

\subsection{Baselines}
\begin{itemize}
    \item \textbf{INT}: This is a vision-based and data-driven approach for robot navigation, named intention-net\cite{gao2017intention}.
    \item \textbf{LiNet}: This method is based on \cite{Pfeiffer2017FromPT}, where the camera sensor of INT is replaced with a lidar to evaluate if OGMs from laser data can bring better performance.
    \item \textbf{UA-LiNet}: This is the state-of-the-art (SOTA) uncertainty-aware lidar-based end-to-end driving method introduced in \cite{liu2021efficient}. Besides the imitation learning objective, the network is also trained to estimate the model's uncertainties, through which the network predictions are dynamically weighted in a sliding window to increase robustness.
    \item \textbf{RL-Lead}: This is the ablated version of our method, where we remove the contrastive loss during training to measure its effects on training efficiency and task performance.
    \item \textbf{FSM-TTC}: This is a traditional rule-based method for crossing intersections referring to \cite{cosgun2017towards}. It is implemented by a finite state machine with the time-to-collision safety indicator\cite{minderhoud2001extended}.
    
\end{itemize}

\begin{table*}[t]
\newcommand{\tabincell}[2]{\begin{tabular}{@{}#1@{}}#2\end{tabular}}
\newcommand{\NA}{---}
        \renewcommand{\arraystretch}{1.3}
        \definecolor{minigray}{rgb}{0.92, 0.92, 0.92}
        \caption{Closed-loop Evaluation Results of Different Models on Four Scenarios in CARLA. $\uparrow$ Means Larger Numbers Are Better, $\downarrow$ Means Smaller Numbers Are Better. The Bold Font Highlights the Best Results in Each Column.}
        \label{tab:evaluation}
        \centering
        \begin{tabular}{l C{1.3cm} C{1.6cm} C{1.3cm} C{1.6cm} C{1.3cm} C{1.6cm} C{1.3cm} C{1.6cm} }
        
        \toprule
        {}&
        \multicolumn{2}{c}{\texttt{T-Left}} &
        \multicolumn{2}{c}{\texttt{T-Merge}} & 
        \multicolumn{2}{c}{\texttt{Int-Cross}} & 
        \multicolumn{2}{c}{\texttt{Int-Left}} \\
        
        \cmidrule(lr){2-3} \cmidrule(lr){4-5} \cmidrule(lr){6-7} \cmidrule(lr){8-9}
        Models & Succ. Rate   & Compl. Time  & Succ. Rate  & Compl. Time & Succ. Rate  & Compl. Time & Succ. Rate & Compl. Time  \\
        \hline
        {\textit{regular traffic}} & (\%) $\uparrow$ & ($s$) $\downarrow$ & (\%) $\uparrow$ & ($s$) $\downarrow$ & (\%) $\uparrow$ & ($s$) $\downarrow$ & (\%) $\uparrow$ & ($s$) $\downarrow$ \\
        \hline
        
        INT\cite{gao2017intention} & 50.50 & 11.05 & 74.50 & 7.50 & 60.50 & 10.15 & 65.00 & 9.97 \\
        LiNet\cite{Pfeiffer2017FromPT} & 85.50 & 9.59 & 93.00 & 6.96 & 91.00 & 8.70 & 83.00 & 8.63 \\
        UA-LiNet\cite{liu2021efficient} & 86.00 & 9.99 & 88.00 & 7.44 & 93.50 & 9.86 & 89.00 & 9.51 \\
        FSM-TTC\cite{cosgun2017towards} & 88.50 & 9.95 & \textbf{98.50} & 9.30 & \textbf{96.50} & 9.11 & \textbf{95.50} & 9.12 \\
        \rowcolor{minigray}
        Carl-Lead \textit{(ours)} & \textbf{93.00} & \textbf{8.49} & 95.00 & \textbf{6.31} & 91.50 & \textbf{7.65} & 93.50 & \textbf{7.97} \\
        RL-Lead & 64.00 & 13.31 & 79.50 & 9.85 & 67.00 & 9.80 & 76.00 & 13.89 \\

        \hline
        \multicolumn{2}{l}{\textit{dense traffic}}\\
        \hline
        
        INT\cite{gao2017intention} & 52.50 & 11.14 & 74.50 & 7.46 & 69.50 & 10.26 & 62.00 & 10.07 \\
        LiNet\cite{Pfeiffer2017FromPT} & 78.00 & 9.60 & 92.00 & 7.08 & 83.50 & 9.67 & 70.50 & 9.16 \\
        UA-LiNet\cite{liu2021efficient} & 83.00 & 10.89 & 85.00 & 7.53 & 88.00 & 10.38 & 82.00 & 9.63 \\
        FSM-TTC\cite{cosgun2017towards} & 82.50 & 11.85 & \textbf{99.00} & 12.32 & \textbf{95.00} & 9.83 & 90.00 & 10.78 \\
        \rowcolor{minigray}
        Carl-Lead \textit{(ours)} & \textbf{88.00} & \textbf{8.80} & 95.50 & \textbf{6.39} & 92.50 & \textbf{7.78} & \textbf{90.50} & \textbf{8.33} \\
        RL-Lead & 51.00 & 12.84 & 80.00 & 9.77 & 69.50 & 9.55 & 64.50 & 13.87 \\
        
        \bottomrule
        \end{tabular}
        \vspace{-0.2cm}
\end{table*}

Note that for the imitation learning-based methods, containing INT, LiNet, and UA-Linet, we collect 600 episodes of expert demonstrations for training in the same scenarios as our method (Sec. \ref{subsec: scenarios}). All these methods use the same backbone ResNet18 as ours, and 3 consecutive frames (containing raw sensor data and map information) as observations for fair comparison. The FSM-TTC method directly uses the ground-truth information of non-occluded objects, such as the locations and speeds of the surrounding vehicles, for navigating the ego-vehicle. We carefully tuned its parameters for safe driving in the four training scenarios. 

\subsection{Evaluation Methods}
\textbf{Benchmark.} We consider two levels of evaluation methods with increasing difficulty. First, in Sec. \ref{subsec: evaluation_seen}, we quantitatively and qualitatively evaluate different driving models in the four seen environments (e.g., \texttt{Int-Left}). However, \textit{200 new random seeds} different from the training setup are used to simulate the traffic. To further increase the diversity of traffic, different methods are tested with two traffic densities with increasing difficulties: \textit{regular} and \textit{dense}.

Second, in Sec. \ref{subsec: evaluation_unseen}, similar to the above pipeline, we evaluate different methods in two unseen dynamic traffic scenarios to examine their generalization capabilities. Specifically, in \texttt{Roundabout}, the ego-vehicle should drive through a two-lane roundabout with four exits and entrances. In \texttt{Five-Way}, the vehicle should take an unprotected left turn at a narrow five-way intersection.

\textbf{Metrics.} We conduct 200 tests for each round of evaluation (6 methods, 6 scenarios, 2 traffic densities, leading to 14,400 episodes in total), and report their average driving performance in two aspects: 1) \textit{Success rate}: An episode is considered to be successful if the agent reaches a certain goal without any collision; 2) \textit{Completion time}: The average time cost for the successful trials.

\begin{figure*}[t]
        \centering
        \includegraphics[width = 2\columnwidth]{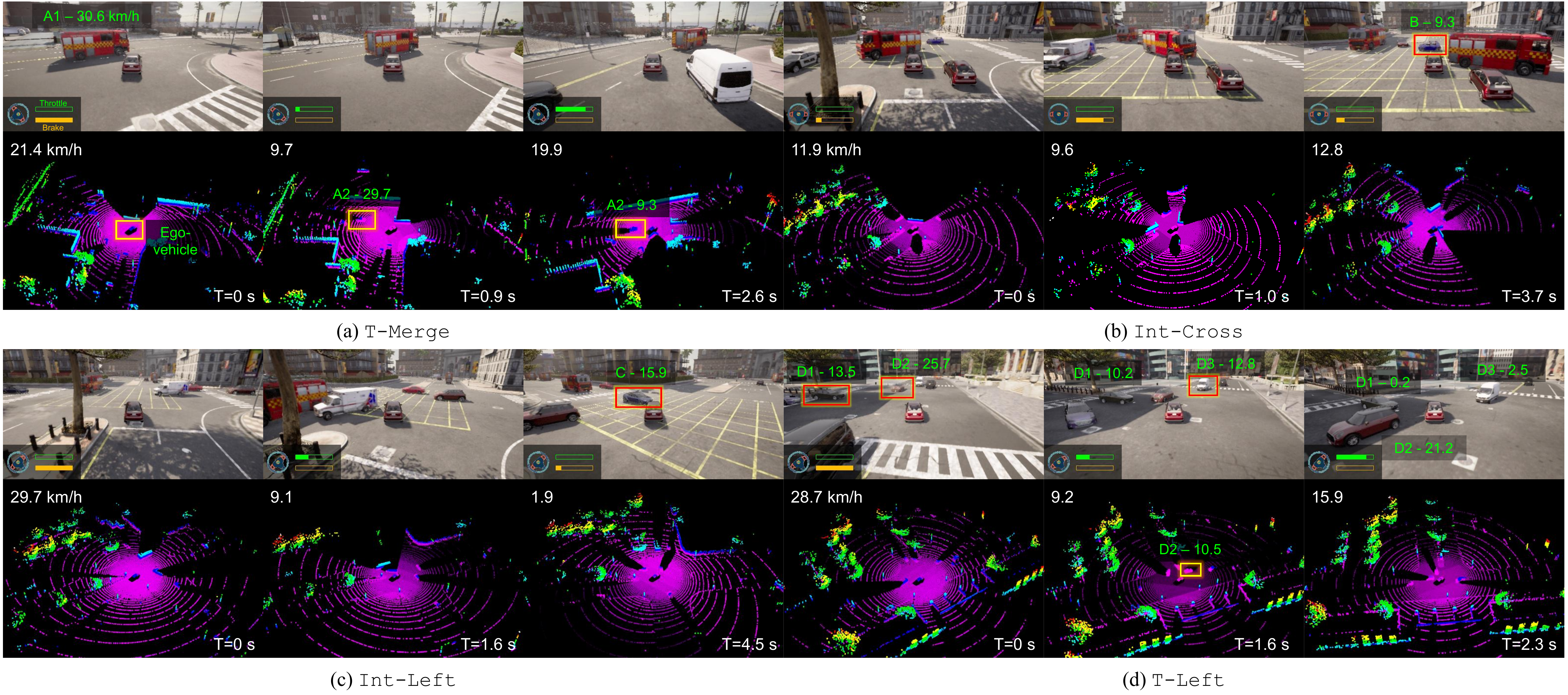}
        \caption{Qualitative evaluation results of Carl-Lead in different driving environments. The ego-speed (in $km/h$) and time stamps are shown in while text. Noticeable other vehicles are bounded by rectangles, and labeled with green texts indicating their ID and speed (in $km/h$). For each scenario, such as (a) \texttt{T-Merge}, the top row shows the third-person views of the environment, and the bottom row shows the lidar point cloud data to help understand the uncertainties and noises of the environment. The sample driving behaviors are: (a) Prepare-for-merge behaviors like humans, (b-c) cautious occlision-handling behaviors, and (d) dynamically-changeable behavioral patterns. Note that the speed information of vehicles is not available to the agent vehicle, and is just annotated for the behavioral analysis. More demonstration videos are available at \url{https://caipeide.github.io/carl-lead/}.}
        \label{fig:qualitative}
\end{figure*}

\begin{figure}[t]
        \centering
        \includegraphics[width = \columnwidth]{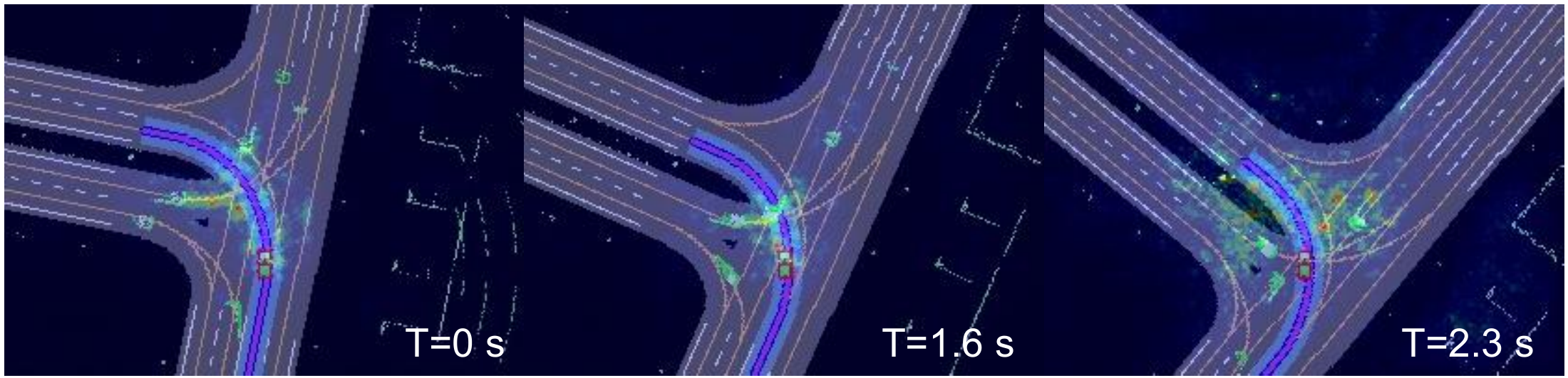}
        \caption{Saliency map showing the attention distribution of the our network model. This is the driving case presented in Fig. \ref{fig:qualitative}-(d) for unprotected left turn at the T-shaped intersection with many other vehicles.}
        \label{fig:saliency}
\end{figure}

\section{Experimental Results}
\label{sec: experiments}
\subsection{Evaluation in Seen Environments\protect\footnote{During evaluation, new seeds are used for the random initialization of traffic participants. Therefore, the environmental dynamics of seen maps is still different from the training period.}}
\label{subsec: evaluation_seen}

\textbf{Quantitative Analysis.} The results are shown in Table \ref{tab:evaluation}. We have the following main findings: 1) Lidar-based networks produce safer results than the vision-based network in our task. For example, the success rate of LiNet is 85.5\% in \texttt{T-Left} with regular traffic, which is about 1.7$\times$ higher than that of INT (50.5\%). We induce that compared with the 2D pixels from cameras, the 3D spatial information from the lidar can help the network learn safer policies. 2) After iterative tuning by designers, the rule-based method FSM-TTC achieves the safest results in general, but may be overly conservative in some scenarios. For example, in \texttt{T-Merge} with dense traffic, FSM-TTC achieves the highest success rate of 99\%, but requires 12.32 s to finish the task, performing worst in terms of driving efficiency. This is because this method controls the vehicle in an \textit{reactive} manner and always waiting for a gap to merge lanes. 3) Our method Carl-Lead achieves close to or higher success rates than the heuristic FSM-TTC method, with the shortest completion time in all environments. For example, in \texttt{T-Left} with regular traffic, the success rate and completion time of our method are 93\% and 8.49 s, respectively, better than those of FSM-TTC (88.5\%, 9.95 s) and other baselines. We conjecture that the DRL training pipeline can help trade off safety and efficiency through trial-and-error in a data-driven manner. 4) The adopted unsupervised contrastive learning is beneficial for improving the task performance, as our method achieves much better results than the ablated model RL-Lead in all environments. In addition, RL-Lead costs 677.7K interaction steps during training to converge, which is about 1.4$\times$ than our method. This reveals that the contrastive learning also leads to higher sample efficiency.

\textbf{Qualitative Analysis.}
\label{subsec: seen-qualitative}
For a more intuitive understanding of our method, we show some qualitative results in Fig. \ref{fig:qualitative} with three types of agent behaviors as follows:

1) \textit{Prepare-for-merge}: In Fig. \ref{fig:qualitative}-(a), the agent vehicle is attempting to merge to the right lane. However, at $t=0$ s, the dangerous red truck A1 blocks the way at a high speed of 30.6 km/h. To ensure safety, the agent applies full brake and slows down for collision avoidance. At $t=0.9$ s, another vehicle A2 approaches and follows the truck at 29.7 km/h. Rather than stop there and give way to A2, the agent applies a prepare-for-merge action similar to humans: it slowly moves forward to the target lane with low throttles, behaving safely and predictably. Then at $t=2.6$ s, we find that A2 responds accordingly and slows down to 9.3 km/h for the agent, which then accelerates to 19.9 km/h for efficient driving.

2) \textit{Occlusion handling}: In Fig. \ref{fig:qualitative}-(b), two big red trucks successively block the front view of the agent, which then slows down and creeps forward cautiously to gain more visibility and ensure safety. At $t=3.7$ s, a blue car B suddenly drives out from the occluded area of the red truck. Since the agent drives slowly in advance, it has enough time to slow down to avoid collision with B. Another similar case is shown in Fig. \ref{fig:qualitative}-(c). At $t=4.5$ s, a black car C drives out from the occluded area of the white truck at a relatively high speed of 15.9 km/h. Since the agent cautiously reacts to the occlusions 4.5 s earlier (slows down from a high speed of 29.7 at $t=0$ s), it has enough time to take small brakes and avoid collision with vehicle C.

3) \textit{Dynamic behavior patterns}: In Fig. \ref{fig:qualitative}-(d), the agent is taking an unprotected left turn at the T-shaped intersection. At $t=0$ s, two vehicles D1 and D2 also drive towards the intersection with unclear intents. Therefore, the agent slows down from 28.7 km/h for safety and a clearer understanding of the environment. At $t=1.6$ s, as D2 continues to drive straight rather than turn right, the vehicle give way to it for collision avoidance. In the mean time, another vehicle D3 approaches the intersection. Rather than stop there, the agent slowly moves forward and take the front lane of D3. Finally, at $t=2.3$ s, D3 understands the intent of the agent and slows down to 2.5 km/h to give way to it. In addition, since D1 stops for D2 in previous frames, it is not dangerous to the agent. Therefore, the agent closely follows the read end of D2 and accelerates to take turns.

Note that the speed information mentioned in the previous analysis is not available to the agent, and we conjecture the agent can infer the motions and intents of other vehicles implicitly through the sequential input observations introduced in Sec. \ref{subsec: state-space}. These cases demonstrate that our Carl-Lead can make the agent \textit{proactively} interactive with the surrounding vehicles, and dynamically adjust its behaviors against uncertainties and occlusions for safe and efficient driving. This also explains the superiority of our method over the \textit{reactive} FSM-TTC in the previous quantitative analysis. 

\begin{table}[t]
\newcommand{\tabincell}[2]{\begin{tabular}{@{}#1@{}}#2\end{tabular}}
\newcommand{\NA}{---}
        \renewcommand{\arraystretch}{1.3}
        \definecolor{minigray}{rgb}{0.92, 0.92, 0.92}
        \caption{Closed-loop Evaluation Results of Different Models on Two Unseen Scenarios in CARLA. $\uparrow$ Means Larger Numbers Are Better, $\downarrow$ Means Smaller Numbers Are Better. The Bold Font Highlights the Best Results in Each Column.}
        \label{tab:generalization}
        \centering
        \begin{tabular}{l C{1.1cm} C{1.1cm} C{1.1cm} C{1.1cm}}
        
        \toprule
        {}&
        \multicolumn{2}{c}{\texttt{Five-Way-Int}} &
        \multicolumn{2}{c}{\texttt{Roundabout}} \\
        
        \cmidrule(lr){2-3} \cmidrule(lr){4-5} 
        Models & Succ. Rt. & C. Time  & Succ. Rt. & C. Time  \\
        \hline
        {\textit{regular traffic}} & (\%) $\uparrow$ & ($s$) $\downarrow$& (\%) $\uparrow$ & ($s$) $\downarrow$ \\
        \hline
        
        INT\cite{gao2017intention} & 74.00 & 8.44 & 4.00 & \textbf{24.62} \\
        LiNet\cite{Pfeiffer2017FromPT} & 78.00 & \textbf{5.95} & 53.00 & 28.59 \\
        UA-LiNet\cite{liu2021efficient} & 80.00 & 7.77 & 51.50 & 28.96 \\
        FSM-TTC\cite{cosgun2017towards} & 93.00 & 8.91 & 67.00 & 30.11 \\
        \rowcolor{minigray}
        Carl-Lead \textit{(ours)} & \textbf{93.50} & 7.63 & \textbf{85.50} & 25.10 \\
        RL-Lead & 82.50 & 20.46 & 69.50 & 42.16 \\

        \hline
        \multicolumn{2}{l}{\textit{dense traffic}}\\
        \hline
        
        INT\cite{gao2017intention} & 73.50 & 8.46 & 6.50 & \textbf{24.87} \\
        LiNet\cite{Pfeiffer2017FromPT} & 68.50 & \textbf{5.97} & 13.50 & 29.61 \\
        UA-LiNet\cite{liu2021efficient} & 76.00 & 8.03 & 11.50 & 30.71 \\
        FSM-TTC\cite{cosgun2017towards} & 89.50 & 12.06 & 35.00 & 40.81 \\
        \rowcolor{minigray}
        Carl-Lead \textit{(ours)} & \textbf{93.00} & 8.23 & \textbf{68.50} & 32.53 \\
        RL-Lead & 66.50 & 20.75 & 30.50 & 44.04 \\

        \bottomrule
        \end{tabular}
        \vspace{-0.2cm}
\end{table}

\subsection{Generalization to New Scenarios}
\label{subsec: evaluation_unseen}
\textbf{Quantitative Analysis}. The generalization results are shown in Table \ref{tab:generalization}. We have the following main findings: 1) In general, all methods encounter performance degradation problems in new scenarios to varying degrees, but the imitation learning-based methods suffer much for this issue. For example, the success rate of INT in \texttt{Roundabout} with regular traffic is only 4\%, which is much lower than its success rates in seen environments (50.5\%$\sim$74.5\%). This is also the case for LiNet and UA-LiNet. By contrast, FSM-TTC has better performance than these methods\footnote{Note that FSM-TTC is provided with perfect perception of the environment for decision making. For real systems, it may suffer from noises and uncertainties from an upper-stream perception module in new environments.}. 2) The carefully tuned heuristic FSM-TTC method generalizes worse than our method in unseen scenarios, although it is provided with perfect sensing information. For example, in \texttt{Roundabout} with dense traffic, its success rate is 35\%, which is about twice lower than ours of 68.5\%.

\begin{figure}[t]
        \centering
        \includegraphics[width = \columnwidth]{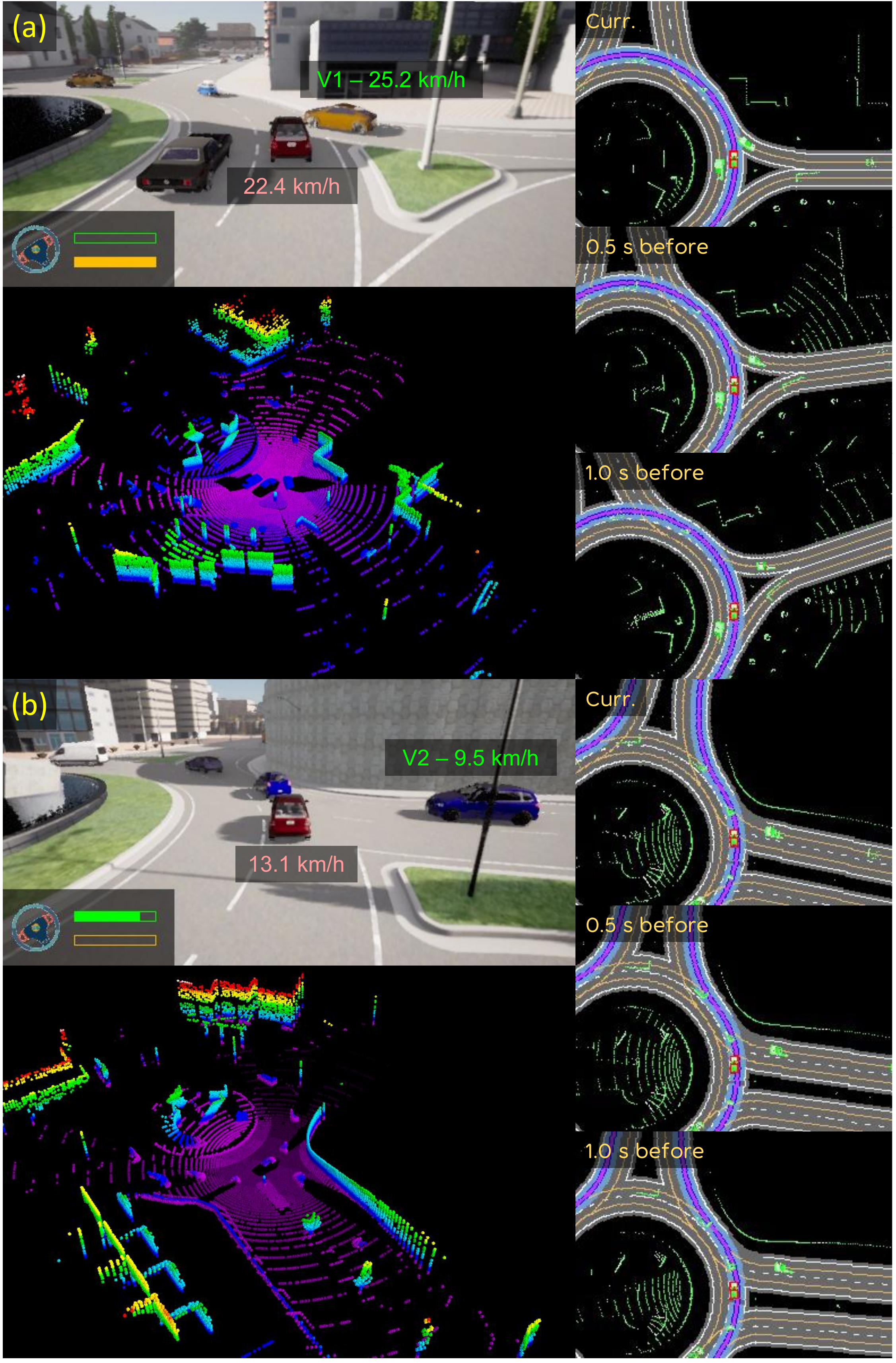}
        \caption{Generalization performance of Carl-Lead on the unseen \texttt{Roundabout} scenario. The presented two driving cases are: (a) Give way to an aggressive vehicle, and (b) take the way of a mild vehicle. The right column shows the sequential observations of the agent.
        }
        \label{fig:generalization}
\end{figure}

\textbf{Qualitative Analysis}.
We show two driving cases in Fig. \ref{fig:generalization} for the unseen \texttt{Roundabout} environment. We can see that the agent takes two different driving decisions on the similar \textit{entrance} environment with other vehicles: slowing down in scenario (a) while accelerating in (b).
In scenario (a), the vehicle V1 is approaching the junction in front of the agent for entering the roundabout. However, V1 is rather aggressive and does not seem to slow down at a high speed of 25.2 km/h. Therefore, the agent timely applies brakes to decelerate for collision avoidance. By contrast, in scenario (b), the vehicle V2 slows down to a speed of 9.5 km/h to give way. Therefore, the agent applies throttles to drive forward. These two cases with opposite agent behaviors further indicate that our method can implicitly infer the intents of other vehicles through the temporal dependency in the observation space (as shown in the right column of Fig. \ref{fig:generalization}), and take actions accordingly.

\subsection{Model Interpretability}
To better understand why better results of our Carl-Lead than other methods can be achieved, in this section we visualize the learned policy by checking the saliency maps\cite{wang2016dueling} of the input observations to see which parts of the environment contribute to the model predictions. Specifically, we compute the absolute value of the Jacobian of the maximum estimated Q value with respect to the input observation: $|\nabla_{s}Q(s, \arg \max_{a^{\prime}}Q(s, a^{\prime}); \theta)|$. 

For example, for the scenario (d)-\texttt{T-Left} of Fig. \ref{fig:qualitative} introduced in the previous qualitative analysis (Sec. \ref{subsec: seen-qualitative}), we show its saliency maps for the input observations in Fig. \ref{fig:saliency}. For the first two time stamps: $t=0$ s and $t=1.6$ s, since the surrounding vehicles exerts a strong influence on the ego-vehicle (which applies brakes and moves forward slowly for collision avoidance), the most salient parts in the observation space are focused around the surrounding vehicles. At $t=2.3$ s, since no other vehicles are dangerous to the ego-vehicle (which is accelerating to turn left), the salient parts are more scattered around the future routes. This phenomenon indicates that our DRL-driven agent can learn to reasonably pay attention to the surroundings according to the dynamically changing driving contexts.

\section{Conclusion}
In this work, to achieve safe and efficient autonomous driving at uncontrolled urban intersections, as well as considering imperfect perceptions such as occlusions, we proposed a lidar-based end-to-end autonomous driving network, named Carl-Lead, where raw lidar data is used to form the partial observation of the environment. We used D3QN as the core DRL algorithm to train the network, combined with parameterized noisy layers for better exploration. During this period, the driving knowledge was learned automatically through trial-and-error without human labels. Moreover, an auxiliary contrastive unsupervised learning branch was designed to further improve the training efficiency and task performance. 

We conducted extensive experiments on four intersection-handling tasks with other dynamic vehicles (e.g., unprotected left turns and lane-merging). The main findings are threefold: 1) The \textit{qualitative} evaluation results showed that our method can learn comprehensive driving skills (e.g., prepare-for-merge, collision avoidance, taking/giving ways and cautiously handling occlusions), and dynamically adjust its behavior patterns in different interactive scenarios according to specific driving contexts. 2) The \textit{quantitative} results showed that our method achieves better results than the supervised imitation learning baselines, and achieves success rates close to or higher than a traditional rule-based method but with less completion time. 3) Furthermore, our method also presents better \textit{generalization} performance than these baselines in new environments (e.g., roundabouts) in terms of safety and efficiency. In the future, we will investigate how to achieve better driving performance of our method, especially in new scenarios, by incorporating safety guarantees or some prior information about the environmental model into the RL learning process.

\bibliographystyle{IEEEtran}
\bibliography{root.bib}

\end{document}